\documentclass{article}
\usepackage[preprint,nonatbib]{style}
\usepackage[utf8]{inputenc}
\usepackage[T1]{fontenc}
\usepackage{hyperref}
\usepackage{url}
\usepackage{booktabs}
\usepackage{amsfonts}
\usepackage{nicefrac}
\usepackage{microtype}
\usepackage{xcolor}
\usepackage[pdftex]{graphicx}
\usepackage{tcolorbox}
\tcbuselibrary{listings, breakable}

\workshoptitle{AI4Mat Workshop}
\title{MatPROV: A Provenance Graph Dataset of Material Synthesis Extracted from Scientific Literature}

\author{
  Hirofumi Tsuruta$^{1}$,~Masaya Kumagai$^{1,2}$\\
  $^1$SAKURA internet Inc.,~$^2$Kyoto University\\
  \texttt{\{hi-tsuruta, m-kumagai\}@sakura.ad.jp}
}

\begin{document}

\maketitle

\begin{abstract}
Synthesis procedures play a critical role in materials research, as they directly affect material properties.
With data-driven approaches increasingly accelerating materials discovery, there is growing interest in extracting synthesis procedures from scientific literature as structured data.
However, existing studies often rely on rigid, domain-specific schemas with predefined fields for structuring synthesis procedures or assume that synthesis procedures are linear sequences of operations, which limits their ability to capture the structural complexity of real-world procedures.
To address these limitations, we adopt PROV-DM, an international standard for provenance information, which supports flexible, graph-based modeling of procedures.
We present MatPROV, a dataset of PROV-DM-compliant synthesis procedures extracted from scientific literature using large language models.
MatPROV captures structural complexities and causal relationships among materials, operations, and conditions through visually intuitive directed graphs.
This representation enables machine-interpretable synthesis knowledge, opening opportunities for future research such as automated synthesis planning and optimization.
\end{abstract}

\section{Introduction}
\label{sec:introduction}

The vast majority of knowledge in materials science exists as unstructured text within the scientific literature.
As data-driven approaches play an increasingly vital role in accelerating materials discovery, substantial efforts have been devoted to extracting structured information from these textual sources~\cite{schilling2025text,jiang2025applications}.
Among the various types of information present in the materials science literature, synthesis procedures represent particularly critical knowledge because they directly affect the resulting material properties~\cite{baig2021nanomaterials,shaba2021critical}.
The systematic extraction and structuring of synthesis procedures hold significant promise for numerous applications, including automated synthesis planning, process optimization, and insights into relationships between synthesis conditions and material properties.

Consequently, numerous studies have extracted and compiled synthesis procedures from scientific literature into publicly available datasets covering various material classes, including metal-organic frameworks (MOFs)~\cite{zheng2023chatgpt,shi2024llm}, gold nanorods~\cite{walker2023extracting}, and a wide range of inorganic materials~\cite{kononova2019text,wang2022dataset}.
Despite these promising developments, existing datasets exhibit notable limitations in how they represent synthesis procedures.
Several datasets~\cite{walker2023extracting,zheng2023chatgpt,shi2024llm} rely on fixed schemas that predefine specific fields.
For example, Shi \textit{et al.}~\cite{shi2024llm} employed a schema specifically designed for MOFs, in which key fields such as ``Metal\_Source,'' ``Organic\_Linker,'' and ``Reaction\_Time'' are predefined in JSON format.
Such domain-specific schemas hinder the development of synthesis knowledge that can be flexibly and generically applied across broader material domains.
In contrast, Kononova \textit{et al.}~\cite{kononova2019text} and Wang \textit{et al.}~\cite{wang2022dataset} proposed representing synthesis procedures as custom-designed ordered sequences of operations, which accommodate arbitrary operations.
However, these approaches assume a linear sequence of synthetic operations from the precursors to the target materials.
This assumption limits their ability to capture the structural complexity commonly encountered in real-world syntheses, such as branching and converging procedures involving multiple synthetic routes.

To address these limitations, we adopt the PROV Data Model (PROV-DM)~\cite{belhajjame2013prov}, an international standard for provenance information established by the World Wide Web Consortium, which enables flexible, graph-based modeling of synthesis procedures.
We present MatPROV, a dataset of PROV-DM-compliant synthesis procedures extracted from scientific papers using large language models (LLMs).
MatPROV captures causal relationships among materials, operations, and conditions, as well as structural complexities such as branching and convergence, through visually intuitive directed graphs.
By adopting this standardized data model, MatPROV ensures interoperability and extensibility, providing a sustainable foundation for machine-interpretable synthesis knowledge that can be leveraged in AI-driven materials discovery.
The main contributions are summarized as follows.

\begin{itemize}
  \item We propose a novel representation framework for synthesis procedures, based on PROV-DM, that models complex procedural structures using directed graphs.
  \item We release MatPROV, a dataset comprising 2,367 synthesis procedures extracted from 1,568 open-access scientific papers, available at \url{https://huggingface.co/datasets/MatPROV-project/MatPROV}.
  \item We conduct experiments with expert-annotated ground truth data to evaluate extraction accuracy, demonstrating the effectiveness of the LLM-based approach in converting unstructured synthesis text into PROV-DM-compliant procedures.
\end{itemize}

\section{MatPROV: A Provenance Graph Dataset of Material Synthesis}

\subsection{Data Representation Framework}
\label{sec:data_representation_framework}

\begin{figure}[t]
  \centering
  \includegraphics[width=\textwidth]{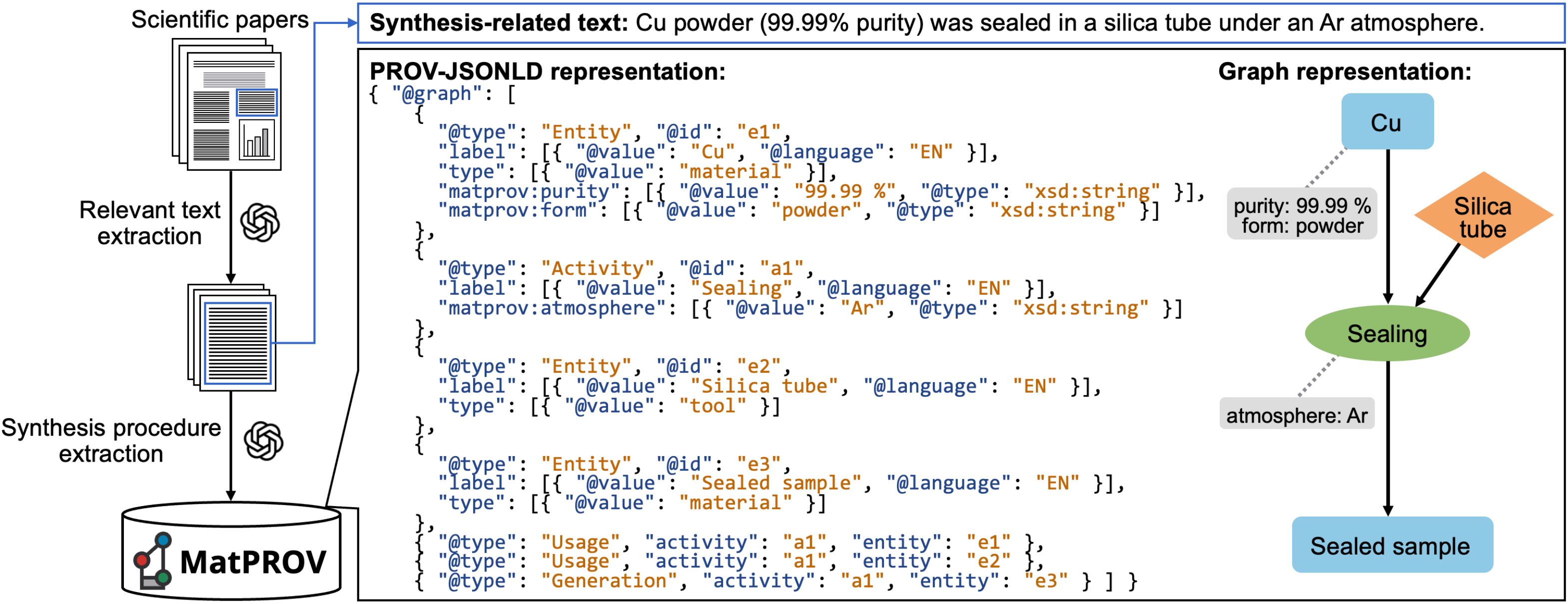}
  \caption{Overview of the dataset construction pipeline and its data representation.
  The example shows a simplified procedure; actual cases typically involve more complex, multi-step processes.}
  \label{fig:matprov_overview}
\end{figure}

The core structure of PROV-DM consists of three types of nodes: entity, activity, and agent.
Provenance is expressed through the use and production of entities by activities, for which agents bear responsibility.
To adapt PROV-DM for representing synthesis procedures, we map entities into two categories: (1) materials, such as precursors, intermediate products, or final products, and (2) experimental tools.
Activities are mapped to experimental operations.
We omit agents because synthesis procedures in scientific papers implicitly assume the experimenter or their affiliated institution as the responsible agent, making this information redundant for our purposes.
We adopt PROV-JSONLD~\cite{huynh2016prov,moreau2020prov}, which enables explicit semantic representation, as the serialization format.
A PROV-JSONLD example, along with its corresponding graph representation, is presented in Figure~\ref{fig:matprov_overview}.

In materials synthesis, not only the materials, tools, and operations but also the synthesis conditions are critical in determining the outcome.
To capture this information, MatPROV extends the PROV-DM framework by associating each node in the provenance graph with ten synthesis parameters---temperature, duration, pressure, mass, length, purity, concentration, rotation, atmosphere, and form---in a manner fully compliant with the PROV-DM specification.
Through this systematic mapping and parameter design, MatPROV leverages the standardized PROV-DM framework to represent synthesis procedures as directed graphs, enabling semantic interoperability and systematic comparison of procedures, and providing significant advantages over ad-hoc representations.

\subsection{Dataset Construction}

\paragraph{Paper Collection}
We utilized Starrydata2~\cite{katsura2019data,katsura2025starrydata}, a web-based database that curates experimental material properties extracted from figures in scientific papers, under the CC BY 4.0 license to select target papers.
Starrydata2 primarily curates property curves for functional inorganic materials, such as temperature-dependent electrical conductivity and Seebeck coefficients, which will support future understanding of property-synthesis relationships using MatPROV.
As Starrydata2 primarily focuses on functional inorganic materials, this source selection introduces a coverage bias toward thermoelectric and magnetic materials, as discussed in Appendix~\ref{sec:appendix_dataset_analysis}.
To ensure copyright compliance, we restricted our collection to open-access publications, resulting in a corpus of 1,648 papers.

\paragraph{Relevant Text Extraction}
Given the extensive volume of text in scientific papers and the computational cost of processing full texts, we extracted only text relevant to synthesis procedures.
First, we converted the downloaded PDF files into structured XML using GROBID~\cite{GROBID} v0.8.2 under the Apache License 2.0 and extracted the main body text while excluding sections such as the title, abstract, and references.
We then employed OpenAI's GPT-4o mini (2024-07-18) with the prompts provided in Appendix~\ref{sec:appendix_relevant_text_extraction} to identify and extract synthesis-related text.
As a result, 32 papers contained no relevant synthesis information, yielding synthesis-relevant text from 1,616 papers.

\paragraph{Synthesis Procedure Extraction}
We extracted synthesis procedures from each relevant text using the OpenAI API with prompts designed to ensure output compliance with the PROV-JSONLD schema.
The prompts were carefully constructed to guide the model in generating connected directed graphs representing synthesis provenance by clarifying the PROV-JSONLD schema, particularly the concepts of nodes (entities and activities), edges (usage and generation), and their interconnections, as detailed in Appendix~\ref{sec:appendix_procedure_extraction}.
The choice of OpenAI model and the in-context examples used in the prompts were determined based on empirical evaluations, as discussed in Section~\ref{sec:experiments}.

\section{Experiments}
\label{sec:experiments}

\subsection{Experimental Settings}
\label{sec:experimental_settings}

To evaluate the extraction performance of LLMs, we randomly sampled 30 papers containing a total of 44 synthesis procedures.
After manually verifying that the relevant text extraction successfully captured all synthesis-related text, a single domain expert created ground truth in PROV-JSONLD format.
Of these papers, 25 were randomly selected as the test set for evaluation, while the remaining five were used as in-context examples.
We selected three recent and cost-effective OpenAI models available as of July 2025: GPT-4.1 mini (2025-04-14), GPT-4.1 (2025-04-14), and o4-mini (2025-04-16).
Evaluation metrics included the collection rate, to assess whether synthesis procedures described in each paper were comprehensively extracted, and extraction performance measured using precision, recall, and F1-score at two levels: (1) the structural level, evaluating the correctness of nodes and edges, and (2) the parametric level, assessing the attributes of correctly extracted nodes.
All results are averaged across five runs with standard deviations.

\begin{table}
  \caption{Performance comparison of different OpenAI models under zero-shot prompting.}
  \centering
  \label{tab:model_comparison_result}
  \resizebox{\textwidth}{!}{
  \begin{tabular}{ccccccccc}
    \toprule
     && \multicolumn{3}{c}{Structural level} && \multicolumn{3}{c}{Parametric level} \\ \cline{3-5} \cline{7-9}
    Model & Collection rate & Precision & Recall & F1-score && Precision & Recall & F1-score \\
    \midrule
    GPT-4.1 mini & 0.724~$\pm$~0.035 & 0.628~$\pm$~0.013 & 0.584~$\pm$~0.024 & 0.605~$\pm$~0.013 && 0.648~$\pm$~0.005 & \textbf{0.754~$\pm$~0.010} & 0.697~$\pm$~0.005 \\
    GPT-4.1 & \textbf{0.930~$\pm$~0.024} & 0.791~$\pm$~0.013 & 0.623~$\pm$~0.022 & 0.697~$\pm$~0.010 && 0.618~$\pm$~0.011 & 0.574~$\pm$~0.019 & 0.595~$\pm$~0.013 \\
    o4-mini & 0.832~$\pm$~0.059 & \textbf{0.792~$\pm$~0.035} & \textbf{0.750~$\pm$~0.031} & \textbf{0.771~$\pm$~0.030} && \textbf{0.766~$\pm$~0.051} & 0.731~$\pm$~0.056 & \textbf{0.748~$\pm$~0.053} \\
    \bottomrule
  \end{tabular}
  }
\end{table}

\begin{table}
  \caption{Performance of o4-mini under one-shot prompting with varying in-context examples.}
  \centering
  \label{tab:one_shot_result}
  \resizebox{\textwidth}{!}{
  \begin{tabular}{cccccccccc}
    \toprule
     && & \multicolumn{3}{c}{Structural level} && \multicolumn{3}{c}{Parametric level} \\ \cline{4-6} \cline{8-10}
    Example DOI & \#Nodes & Collection rate & Precision & Recall & F1-score && Precision & Recall & F1-score \\
    \midrule
    10.1002/advs.201901598~\cite{mao2020decoupling} & 20 & \textbf{0.881~$\pm$~0.049} & \textbf{0.855~$\pm$~0.038} & \textbf{0.816~$\pm$~0.019} & \textbf{0.835~$\pm$~0.026} && \textbf{0.844~$\pm$~0.033} & \textbf{0.805~$\pm$~0.023} & \textbf{0.824~$\pm$~0.022} \\
    10.1038/s41467-019-09921-4~\cite{manley2019intrinsic} & 13 & 0.822~$\pm$~0.049 & 0.783~$\pm$~0.023 & 0.790~$\pm$~0.053 & 0.786~$\pm$~0.034 && 0.802~$\pm$~0.016 & 0.758~$\pm$~0.029 & 0.779~$\pm$~0.013 \\
    10.1063/1.4903773~\cite{wei2014tau} & 9 & 0.870~$\pm$~0.040 & 0.791~$\pm$~0.048 & 0.721~$\pm$~0.024 & 0.754~$\pm$~0.034 && 0.720~$\pm$~0.012 & \textbf{0.805~$\pm$~0.038} & 0.760~$\pm$~0.022 \\
    10.1155/2018/9380573~\cite{boolchandani2018preparation} & 8 & 0.827~$\pm$~0.062 & 0.789~$\pm$~0.032 & 0.747~$\pm$~0.048 & 0.767~$\pm$~0.038 && 0.768~$\pm$~0.021 & 0.760~$\pm$~0.012 & 0.764~$\pm$~0.010 \\
    10.12693/aphyspola.144.333~\cite{przybyl2023magnetic} & 13 & 0.827~$\pm$~0.045 & 0.813~$\pm$~0.030 & 0.732~$\pm$~0.037 & 0.771~$\pm$~0.033 && 0.803~$\pm$~0.042 & 0.766~$\pm$~0.031 & 0.784~$\pm$~0.032 \\
    \bottomrule
  \end{tabular}
  }
\end{table}

\begin{figure}[t]
  \centering
  \includegraphics[width=\textwidth]{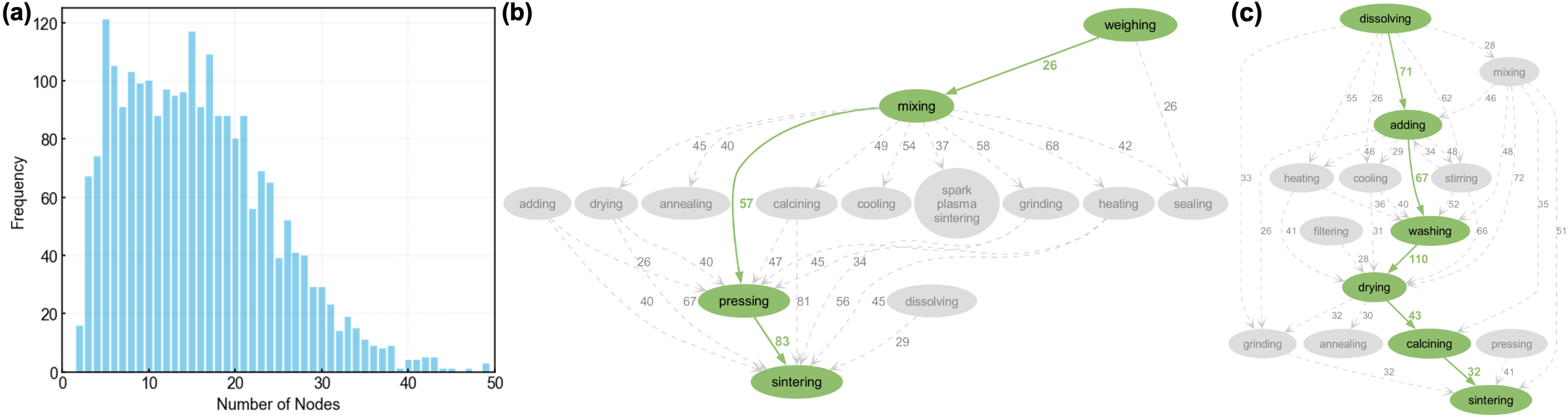}
  \caption{(a) Histogram of node counts per synthesis procedure graph in MatPROV.
  Synthesis backbones for (b) thermoelectric and (c) magnetic materials, represented by green nodes and edge.
  Edge weights represent the co-occurrence frequencies.
  See Appendix~\ref{sec:appendix_dataset_analysis} for details.}
  \label{fig:data_analysis}
\end{figure}

\subsection{Results}

Table~\ref{tab:model_comparison_result} shows the performance comparison of models under zero-shot prompting.
The reasoning-based o4-mini outperformed the general-purpose GPT-4.1 in both structural and parametric-level extraction.
This suggests that extracting step-by-step synthesis procedures as directed graphs requires advanced reasoning capabilities to accurately capture structural relationships and parameters.
Although GPT-4.1 achieved a higher collection rate in identifying synthesis procedures, we prioritized extraction accuracy over coverage in dataset construction, as erroneous records would compromise data quality more than missing several procedures.
We further investigated the effect of in-context examples by focusing on o4-mini. 
Table~\ref{tab:one_shot_result} presents performance under one-shot prompting, where each of five randomly sampled papers was provided as an example.
Performance varied considerably across examples, with several failing to improve upon zero-shot results.
However, the best-performing example yielded F1-score improvements of 6.4\% (structural) and 7.6\% (parametric) over the zero-shot setting.
This variation is likely due to differences in the complexity of synthesis procedures across examples.
In particular, the example with digital object identifier (DOI) ``10.1002/advs.201901598~\cite{mao2020decoupling}'' contained the most complex graph, with 20 nodes compared with other examples, and providing such complex graphs as examples was found to enhance performance.

\section{Dataset Analysis}
\label{sec:dataset_analysis}

We applied one-shot prompting using the DOI ``10.1002/advs.201901598'' with o4-mini to 1,616 papers.
Of these, 48 papers were identified as not containing synthesis procedures, resulting in a dataset of 2,367 synthesis procedures from the remaining 1,568 papers.
Among these, 2,322 procedures (98.1\%) successfully formed directed acyclic graphs, while 4 (0.2\%) resulted in cyclic graphs, and 41 (1.7\%) included isolated nodes.
Figure~\ref{fig:data_analysis}(a) shows a histogram of node counts per graph.
Most graphs contain approximately 5 to 20 nodes, while several exceed 30 nodes, reflecting varying procedural complexity.
Figures~\ref{fig:data_analysis}(b) and (c) illustrate the backbone structures of synthesis procedures for thermoelectric and magnetic materials, extracted using the method described in Appendix~\ref{sec:appendix_dataset_analysis}.
The differences between these material classes reflect actual variations in synthesis methodology.
Thermoelectric materials typically involve powder-based preprocessing steps, such as weighing, mixing, forming, and sintering~\cite{d2023thermoelectric}.
In contrast, magnetic materials often rely on wet-chemistry approaches emphasizing compositional homogeneity through dissolving, adding, washing, and drying~\cite{ali2021review}.
These results indicate that the extracted graphs capture not only procedural steps but also meaningful synthesis patterns, supporting their validity from a materials science perspective.

\section{Limitations and Future Works}
\label{sec:limitations}

MatPROV remains relatively small in scale compared with several existing datasets.
Nevertheless, our LLM-based extraction approach is applicable across the full breadth of scientific literature.
Thus, we plan to expand the scope of source literature, enabling the collection of more diverse synthesis procedures and enhancing the dataset’s scale and generalizability.
Moreover, our evaluation was based on a relatively small test set comprising 37 synthesis procedures from 25 papers, all annotated by a single domain expert.
This limitation prevented us from measuring inter-annotator agreement or quantitatively assessing annotation reliability.
In future work, we plan to involve multiple annotators and report inter-annotator agreement metrics for node, edge, and parameter labeling to better quantify labeling consistency and enhance the robustness of the evaluation framework.
Furthermore, extraction accuracy requires further improvement.
In future work, we aim to improve extraction accuracy through more sophisticated prompt engineering and fine-tuning strategies for LLMs.

%%%%%%%%%%%%%%%%%%%%%%%%%%%%%%%%%%%%%%%%%%%%%%%%%%%%%%%%%%%%

\newpage

\bibliographystyle{splncs04}

\newpage

\appendix

\section{Appendix}

\subsection{Data and Code Availability}
\label{sec:appendix_data_availability}

The MatPROV dataset, comprising 2,367 material synthesis procedures extracted from scientific papers, and the ground truth dataset manually annotated by a domain expert are publicly available under the CC BY 4.0 license at \url{https://huggingface.co/datasets/MatPROV-project/MatPROV}.
The code used for dataset construction and extraction performance evaluations is also publicly available at \url{https://github.com/MatPROV-project/matprov-experiments}.
The JSON-LD context schema defining the vocabulary for material characteristics and synthesis parameters used in the PROV-DM-based provenance graphs is publicly accessible at \url{https://matprov-project.github.io/matprov-schema}.

\subsection{Negative Societal Impacts}
\label{sec:negative_impacts}

Potential risks in MatPROV arise from data quality issues inherent in automated extraction.
The dataset is constructed using LLM-based extraction from scientific papers, which introduces the possibility of extraction errors or misinterpretation of synthesis procedures.
As demonstrated in our experiments, the automated extraction process achieves imperfect accuracy.
If researchers rely solely on the extracted provenance graphs without validating them against the original sources, this could propagate incorrect synthesis information, potentially resulting in failed experiments, resource waste, or, in extreme cases, safety hazards in laboratory settings.
To mitigate these risks, we strongly recommend that users validate critical synthesis procedures against the original papers before practical implementation.
To support this validation, our dataset includes links between extracted synthesis procedures and their corresponding paper DOIs, enabling users to trace back to the original publications easily.

\subsection{Relevant Text Extraction}
\label{sec:appendix_relevant_text_extraction}

We used OpenAI's GPT-4o mini (2024-07-18) with the temperature parameter set to 0.0 for relevant text extraction.
A complete prompt used for extracting synthesis-related text from scientific papers is shown as follows.

\begin{tcolorbox}[
    fontupper=\ttfamily\footnotesize,
    colback=white,
    colframe=black,
    arc=3pt,
    boxrule=0.5pt,
    left=5pt,
    right=5pt,
    top=5pt,
    bottom=5pt,
    breakable
]
\# Task\\
You are a materials science expert. Your task is to extract all paragraphs from the provided "Materials Science Text" that describe material synthesis procedures performed by the authors.\\\\
\# General Instructions\\
-- Output ONLY the exact text of each extracted paragraph, preserving original line breaks and without modifying, splitting, or combining them.\\
-- Do not include any explanations, summaries, or comments in your output.\\
-- If multiple relevant paragraphs are found, include all in their original order.\\
-- If no relevant paragraphs are found, output nothing (leave the response empty).\\\\
\# Extraction Rules\\
-- Do not include paragraphs that only discuss characterization, measurement, analysis, or results unless such content is embedded within a synthesis description.\\
-- Do not extract synthesis procedures if they refer to prior work or the procedures of other researchers (e.g., "Smith et al. synthesized..." or "previous reports prepared...").\\\\
The following categories of synthesis steps should be extracted:\\
-- Raw material preparation and handling:\hspace{0.3em}e.g., sourcing, weighing, drying, or pre-treatment of components.\\
-- Mixing and powder processing:\hspace{0.3em}e.g., manual or mechanical grinding, ball milling, mixing with solvents.\\
-- Forming and compaction:\hspace{0.3em}e.g., pressing, molding, casting, or shaping into desired geometries.\\
-- Thermal treatments:\hspace{0.3em}e.g., annealing, sintering, calcination, with specified temperatures and durations.\\
-- Chemical synthesis steps:\hspace{0.3em}e.g., hydrothermal, solvothermal, precipitation, or solid-state reactions.\\
-- Crystal growth processes:\hspace{0.3em}e.g., flux growth, solution growth, vapor transport, or melt growth, with specified conditions such as temperature gradients, cooling rates, and growth atmospheres.\\
-- Post-processing:\hspace{0.3em}e.g., quenching, cooling, polishing, etching, aging, surface treatments, or irradiating.\\\\
\# Example Paragraph\\
Polycrystalline Cu2-$\delta$FexS ($\delta$ = 0.1, x = 0, 0.0125, 0.0225, and 0.0325) and Cu2-$\delta$S ($\delta$ = 0, 0.01, 0.03, 0.04, 0.06, and 0.1) samples were synthesized by a combination of melting and long-term high-temperature annealing method. High purity raw elements, Cu (shot, 99.999\%, Alfa Aesar), S (shot, 99.999\%, Alfa Aesar), and Fe (shots, 99.98\%, Alfa Aesar) were weighed in their stoichiometric ratios and placed in boron nitride crucibles, and then sealed in fused silica tubes under vacuum. The temperature of the tubes was slowly raised to 1423 K in 6 h and then maintained at this temperature for 12 h before quenching into ice water. Then, the ingots were annealed at 773 K for 5 d. The annealed ingots were crushed into powders and consolidated by spark plasma sintering (Sumitomo SPS-2040) at 723 K under a pressure of 65 MPa for 5 min. Electrically insulating but thermally conducting BN layers were sprayed onto the carbon foils and the inner sides of the graphite die before the SPS process in order to prohibit DC pulsed currents going through the powders.\\\\
\# Materials Science Text\\
<PAPER\_TEXT>
\end{tcolorbox}

The placeholder <PAPER\_TEXT> is dynamically replaced with the body text of each paper.
We manually confirmed that relevant text extraction successfully captured all synthesis-related text from the 25 test set papers used in Section~\ref{sec:experiments}, with no omissions.
This ensured that the subsequent evaluation reflected the models’ ability to extract structured information without being affected by incomplete input.

\subsection{Synthesis Procedure Extraction}
\label{sec:appendix_procedure_extraction}

We used OpenAI's GPT-4.1 mini (2025-04-14) and GPT-4.1 (2025-04-14) with the temperature parameter set to 0.0, and o4-mini (2025-04-16) without a temperature setting (as it is not supported), for synthesis procedure extraction.
A complete prompt used for extracting synthesis procedures in PROV-JSONLD format from synthesis-related text is shown as follows.

\begin{tcolorbox}[
    fontupper=\ttfamily\footnotesize,
    colback=white,
    colframe=black,
    arc=3pt,
    boxrule=0.5pt,
    left=5pt,
    right=5pt,
    top=5pt,
    bottom=5pt,
    breakable
]
\# Task\\
You are a materials science expert. Your task is to extract the material synthesis procedure described in the provided "Materials Science Text" and represent it as a directed acyclic graph (DAG) based on the PROV Data Model (PROV-DM).\\\\
\# General Instructions\\
-- Output ONLY valid JSON. Do NOT include any explanations, comments, or Markdown formatting in your output.\\
-- Extract information exactly as stated in the input text. Do NOT paraphrase, infer, generalize, or modify the original wording.\\
-- The input text may contain paragraphs that are not related to material synthesis. Extract information only from paragraphs that describe actual material synthesis procedures.\\
-- The input text may describe multiple distinct synthesis procedures. If procedures differ in nodes, edges, or node labels (e.g., due to different activity sequences, materials, equipment, or target compositions), extract each as a separate JSON object. If procedures have identical nodes, edges, and node labels but differ only in parameter values, combine them into a single JSON object.\\\\
\# Output JSON Structure\\
Each material synthesis procedure must be represented as a JSON object with exactly two top-level keys:\hspace{0.3em}"label" and "@graph". Do NOT add any additional top-level keys. The JSON structure rules are explained using the following minimal sample.\\\\
\`{}\`{}\`{}json
\begin{verbatim}
[
  {
    "label": "<chemical composition>_<characteristic>",
    "@graph": [
      {
        "@type": "Entity",
        "@id": "e1",
        "label": [{ "@value": "Cu" }],
        "type": [{ "@value": "material" }],
        "matprov:purity": [{ "@value": "99.99 %" }],
        "matprov:form": [{ "@value": "pieces" }]
      },
      {
        "@type": "Activity",
        "@id": "a1",
        "label": [{ "@value": "Sealing" }]
      },
      {
        "@type": "Entity",
        "@id": "e2",
        "label": [{ "@value": "silica tube" }],
        "type": [{ "@value": "tool" }]
      },
      {
        "@type": "Entity",
        "@id": "e3",
        "label": [{ "@value": "Sealed sample" }],
        "type": [{ "@value": "material" }]
      },
      { "@type": "Usage", "activity": "a1", "entity": "e1" },
      { "@type": "Usage", "activity": "a1", "entity": "e2" },
      { "@type": "Generation", "activity": "a1", "entity": "e3" }
    ]
  }
]
\end{verbatim}
\`{}\`{}\`{}\\\\
Each object in "@graph" represents either a node ("@type":\hspace{0.3em}"Activity" or "Entity") or an edge ("@type":\hspace{0.3em}"Usage" or "Generation") in a DAG that describes the provenance chain of the material synthesis procedure. The minimal sample above corresponds to the following graph:\\\\
Cu \& silica tube $\rightarrow$ sealing $\rightarrow$ sealed sample\\\\
IMPORTANT:\hspace{0.3em}All nodes (Activity or Entity) must be connected by at least one edge (Usage or Generation). For each synthesis procedure, construct a single connected graph where every node is reachable from every other node via directed edges, forming a continuous provenance chain. Do not leave any node isolated or disconnected. Avoid creating disconnected subgraphs or unlinked activities/entities. Reuse intermediate @ids appropriately to ensure continuity across multiple synthesis steps.\\\\
\#\# Nodes\\
Fill in the "@value" of "label" for each node following the rules below. Node labels MUST represent single atomic concepts -- NEVER use "and" in labels. Split items joined by "and" into separate nodes.\\\\
\#\#\# Activity\\
Use the gerund form of the verb as the label (e.g., melting, crushing, sealing, adding, ball-milling). Include modifying terms in the label (e.g., spark plasma sintering, arc-melting).\\\\
\#\#\# Entity\\
Entity has two types:\hspace{0.3em}material and tool.\\\\
1. material\\
-- Precursors:\hspace{0.3em}Use the names or symbols exactly as presented in the input text (e.g., element symbols, full names).\\
-- Intermediate/Final products:\hspace{0.3em}MANDATORY RULE:\hspace{0.3em}The label MUST be exactly "<past participle> sample" where the past participle corresponds to the Activity that generates the Entity.\\
\hspace*{1em}-- Examples:\hspace{0.3em}arc-melting $\rightarrow$ arc-melted sample, crushing $\rightarrow$ crushed sample, spark plasma sintering $\rightarrow$ spark plasma sintered sample\\
\hspace*{1em}-- Physical form information (e.g., ingot, powder, pellet) must be recorded in the "matprov:form" parameter, NOT in the label.\\\\
2. tool\\
-- Extract every apparatus and tool as a single generic noun phrase from the text (e.g., graphite die, furnace).\\
-- If model name and company name are both given, exclude the company name (e.g., "ARC-2000 furnace, ABC Corp." $\rightarrow$ "ARC-2000 furnace").\\\\
\#\# Edges\\
Each edge type represents a directed connection between nodes as follows:\\
-- Usage:\hspace{0.3em}Entity $\rightarrow$ Activity\\
-- Generation:\hspace{0.3em}Activity $\rightarrow$ Entity\\\\
Use the following format:\\\\
\`{}\`{}\`{}json
\begin{verbatim}
{ "@type": "Usage", "activity": "<unique id>", "entity": "<unique id>" },
{ "@type": "Generation", "activity": "<unique id>", "entity": "<unique id>" }
\end{verbatim}
\`{}\`{}\`{}\\\\
\#\# Parameters\\
Attach only explicitly stated parameters to relevant nodes using the following format:\\\\
\`{}\`{}\`{}json
\begin{verbatim}
"matprov:<parameter>": [{ "@value": "<value>" }]
\end{verbatim}
\`{}\`{}\`{}\\\\
Accepted Parameters (10):\\
temperature, duration, pressure, mass, length, purity, concentration, rotation, atmosphere, form\\\\
Modifiers (7):\\
-- Global modifiers:\hspace{0.3em}\_start, \_end, \_rate\\
-- Length-specific modifiers:\hspace{0.3em}\_width, \_height, \_thickness, \_diameter\\
\hspace*{1em}-- Example:\hspace{0.3em}"matprov:length\_thickness"\\\\
Parameter placement:\\
-- Activity nodes:\hspace{0.3em}Process conditions (e.g., temperature, duration, pressure, mass, concentration, rotation, atmosphere, form)\\
-- Entity nodes:\hspace{0.3em}Object descriptors (e.g., mass, length, purity, concentration, form)\\\\
IMPORTANT:\hspace{0.3em}If multiple values are mentioned for the same parameter in the input text (e.g., "annealed at 100, 200, and 300 K"), combine all values into a single @value string, preserving the original wording as follows:\\\\
\`{}\`{}\`{}json\\
"matprov:temperature": [{ "@value": "100, 200, and 300 K" }]\\
\`{}\`{}\`{}\\\\
Do NOT output each value as a separate dictionary in the parameter list.\\\\
\# Example\\
<IN\_CONTEXT\_EXAMPLE>\\\\
\# Materials Science Text\\
<SYNTHESIS\_TEXT>
\end{tcolorbox}

The placeholders <SYNTHESIS\_TEXT> and <IN\_CONTEXT\_EXAMPLE> are dynamically replaced with synthesis-related text and a single in-context example.
Common metadata fields in PROV-JSONLD, such as ``@context'' and ``@language'', are automatically appended through rule-based post-processing rather than being generated by the LLM.
This prevents potential formatting errors and reduces unnecessary computational load on the LLM.

\subsection{Evaluation Methodology}
\label{sec:appendix_evaluation}

\paragraph{Procedure Matching and Collection Rate}

To account for the possibility that a single scientific paper may contain multiple distinct synthesis procedures, we prompt the LLM to comprehensively extract all available procedures.
When either the ground truth or the LLM output includes multiple procedures, it is necessary to establish correspondence between individual procedures to form appropriate procedure pairs for evaluating extraction accuracy.
To facilitate this, each extracted synthesis graph includes a ``label'' field---alongside the top-level ``@graph'' key in the JSON format (see the prompt in Appendix~\ref{sec:appendix_procedure_extraction})---which encodes the material's chemical composition and key synthesis characteristics (e.g., ``CuGaTe2\_ball-milling'').
We compute string similarity scores for all possible pairs of ground truth and LLM-generated labels using Python's difflib.SequenceMatcher, and match them iteratively in descending order of similarity.
If the LLM generates more procedures than present in the ground truth, unmatched LLM-generated procedures are excluded from evaluation.
Conversely, if the LLM generates fewer procedures than the ground truth contains, this represents a failure to collect procedures that should have been identified.
To quantify this aspect, we define the collection rate as the ratio of the number of matched LLM-generated procedures to the total number of ground truth procedures, reflecting the model’s ability to comprehensively identify synthesis procedures within a paper.

\paragraph{Performance Metrics}

For the matched procedure pairs previously identified, we evaluate extraction performance using precision, recall, and F1-score. 
Precision is defined as the proportion of elements extracted by the LLM that are actually correct, recall is defined as the proportion of ground truth elements that are correctly extracted by the LLM, and F1-score is the harmonic mean of precision and recall.
The choice of granularity for defining ``elements'' significantly impacts evaluation outcomes.
In this study, we evaluate performance at two levels: (1) the structural level, assessing the correctness of nodes and edges, and (2) the parametric level, treating each key-value pair associated with graph nodes as a separate element.
Parameter evaluation is performed only for nodes that were correctly extracted.

\paragraph{String Matching Criteria}

We describe the criteria used to determine whether individual graph elements—nodes, edges, and parameters—are correctly extracted.
Each node contains a ``label'' field specifying material names, tool names, or similar descriptors (e.g., ``Cu'' or ``silica tube''), which serves as the basis for node-level evaluation.
Each edge includes the unique identifiers of its source and target nodes, enabling retrieval of their node labels for edge evaluation.
Parameters are represented as key-value pairs, with values denoting specific attributes (e.g., ``99.99\%'' for ``matprov:purity''), and both key and value strings are considered during evaluation.
While string matching offers an intuitive basis for evaluating correctness, strict exact matches are often too strict and may incorrectly penalize semantically valid extractions.
To address this, we adopt two strategies to improve robustness.
First, we normalize strings prior to comparison by applying operations such as lowercasing and removing special characters (e.g., hyphens and underscores).
Second, we allow multiple acceptable variants for each ground truth element to account for common linguistic and domain-specific variations.
These variants address ambiguities, including notational alternatives (e.g., ``Ar'' vs. ``Argon''), differing levels of descriptive specificity (e.g., ``ball-milling'' vs. ``mechanical ball-milling''), and variations in linguistic form (e.g., ``calcining'' vs. ``calcination'').
An extracted element is considered correct if its normalized form matches any of the normalized acceptable ground truth variants.
This flexible matching ensures that the evaluation reflects semantic correctness rather than superficial textual differences.

\subsection{Additional Results}

\begin{table}
  \caption{Node-level and edge-level performance comparison of different OpenAI models under zero-shot prompting.}
  \centering
  \label{tab:appendix_model_comparison_result}
  \resizebox{\textwidth}{!}{
  \begin{tabular}{ccccccccc}
    \toprule
     & \multicolumn{3}{c}{Node level} && \multicolumn{3}{c}{Edge level} \\ \cline{2-4} \cline{6-8}
    Model & Precision & Recall & F1-score && Precision & Recall & F1-score \\
    \midrule
    GPT-4.1 mini & 0.727~$\pm$~0.013 & 0.673~$\pm$~0.024 & 0.699~$\pm$~0.011 && 0.525~$\pm$~0.016 & 0.491
~$\pm$~0.027 & 0.507~$\pm$~0.018 \\
    GPT-4.1 & \textbf{0.855~$\pm$~0.009} & 0.669~$\pm$~0.024 & 0.751~$\pm$~0.015 && 0.725~$\pm$~0.019 & 0.575~$\pm$~0.022 & 0.641~$\pm$~0.007 \\
    o4-mini & 0.844~$\pm$~0.024 & \textbf{0.799~$\pm$~0.024} & \textbf{0.821~$\pm$~0.021} && \textbf{0.738~$\pm$~0.046} & \textbf{0.699~$\pm$~0.039} & \textbf{0.718~$\pm$~0.040} \\
    \bottomrule
  \end{tabular}
  }
\end{table}

\begin{table}
  \caption{Node-level and edge-level performance of o4-mini under one-shot prompting with varying in-context examples.}
  \centering
  \label{tab:appendix_one_shot_result}
  \resizebox{\textwidth}{!}{
  \begin{tabular}{cccccccc}
    \toprule
    & \multicolumn{3}{c}{Node level} && \multicolumn{3}{c}{Edge level} \\ \cline{2-4} \cline{6-8}
    Example DOI & Precision & Recall & F1-score && Precision & Recall & F1-score \\
    \midrule
    10.1002/advs.201901598~\cite{mao2020decoupling} & \textbf{0.892~$\pm$~0.023} & \textbf{0.851~$\pm$~0.020} & \textbf{0.871~$\pm$~0.017} && \textbf{0.817~$\pm$~0.058} & \textbf{0.780~$\pm$~0.036} & \textbf{0.798~$\pm$~0.046} \\
    10.1038/s41467-019-09921-4~\cite{manley2019intrinsic} & 0.832~$\pm$~0.014 & 0.834~$\pm$~0.045 & 0.833~$\pm$~0.024 && 0.731~$\pm$~0.034 & 0.744~$\pm$~0.061 & 0.737~$\pm$~0.044 \\
    10.1063/1.4903773~\cite{wei2014tau} & 0.854~$\pm$~0.030 & 0.776~$\pm$~0.020 & 0.813~$\pm$~0.020 && 0.725~$\pm$~0.066 & 0.662~$\pm$~0.037 & 0.692~$\pm$~0.050 \\
    10.1155/2018/9380573~\cite{boolchandani2018preparation} & 0.837~$\pm$~0.027 & 0.790~$\pm$~0.047 & 0.813~$\pm$~0.036 && 0.739~$\pm$~0.037 & 0.702~$\pm$~0.050 & 0.720~$\pm$~0.042 \\
    10.12693/aphyspola.144.333~\cite{przybyl2023magnetic} & 0.858~$\pm$~0.017 & 0.776~$\pm$~0.026 & 0.815~$\pm$~0.022 && 0.766~$\pm$~0.044 & 0.687~$\pm$~0.050 & 0.724~$\pm$~0.047 \\
    \bottomrule
  \end{tabular}
  }
\end{table}

We present a breakdown of structural-level extraction performance for nodes and edges separately, in addition to the overall structural-level extraction performance shown in Tables~\ref{tab:model_comparison_result} and~\ref{tab:one_shot_result}.
Tables~\ref{tab:appendix_model_comparison_result} and~\ref{tab:appendix_one_shot_result} report node-level and edge-level performance metrics under zero-shot and one-shot prompting, respectively.
Edge-level evaluation is inherently dependent on the correct identification of the corresponding nodes.
If a node is not correctly extracted, all edges associated with that node are automatically considered incorrect.
As a result, edge-level performance is often penalized not only by errors in edge prediction but also by upstream node extraction errors.

\begin{figure}[t]
  \centering
  \includegraphics[width=\textwidth]{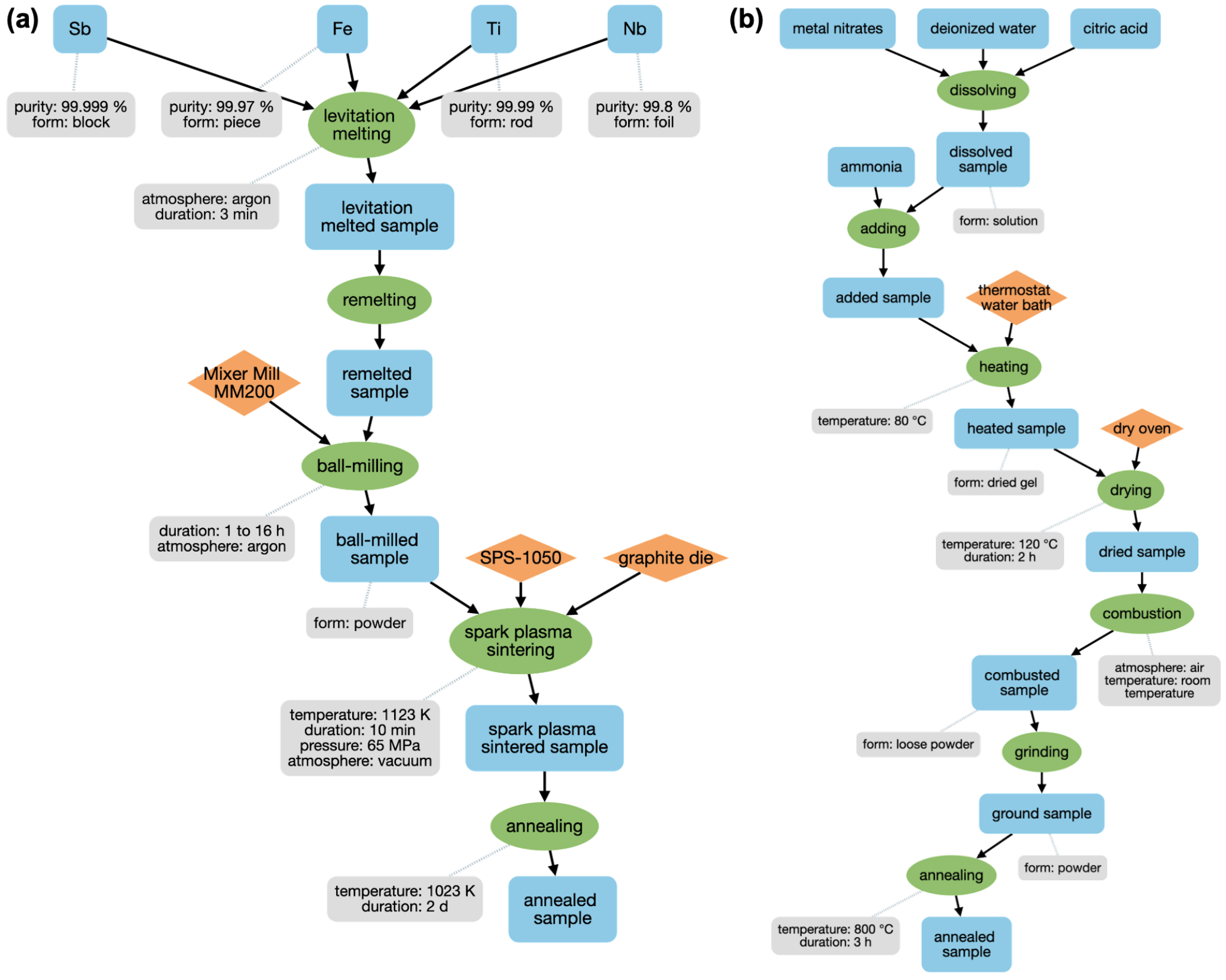}
  \caption{Representative examples of synthesis procedure graphs extracted using o4-mini from the paper with DOI (a) ``10.1002/advs.201600035~\cite{fu2016enhancing}'' and (b) ``10.1155/2015/854840~\cite{he2015mossbauer}.''}
  \label{fig:LLM-generated_graphs}
\end{figure}

To demonstrate the capability of LLMs in extracting complex synthesis procedure graphs, Figure~\ref{fig:LLM-generated_graphs} visualizes extracted graphs for two representative test set papers.
Figure~\ref{fig:LLM-generated_graphs}(a) illustrates the synthesis procedure of a thermoelectric material, extracted from the paper with DOI ``10.1002/advs.201600035~\cite{fu2016enhancing}'', achieving structural-level metrics of recall 0.849, precision 0.966, and F1 score 0.903, and parametric-level metrics of recall 0.900, precision 0.947, and F1 score 0.923.
Figure~\ref{fig:LLM-generated_graphs}(b) illustrates the synthesis of a magnetic material, extracted from the paper with DOI ``10.1155/2015/854840~\cite{he2015mossbauer}'', achieving structural-level metrics of recall 0.861, precision 0.841, and F1 score 0.851, and parametric-level metrics of recall 0.889, precision 1.000, and F1 score 0.941.
Although edges in PROV-DM are defined to represent provenance by pointing backward in time (from result to source), we reverse the arrow directions in our visualizations for clarity.
This aligns the graph layout with the natural temporal flow of experimental procedures, improving readability and interpretability of the extracted synthesis graphs.

\subsection{Dataset Analysis}
\label{sec:appendix_dataset_analysis}

\begin{figure}
  \centering
  \includegraphics[width=\textwidth]{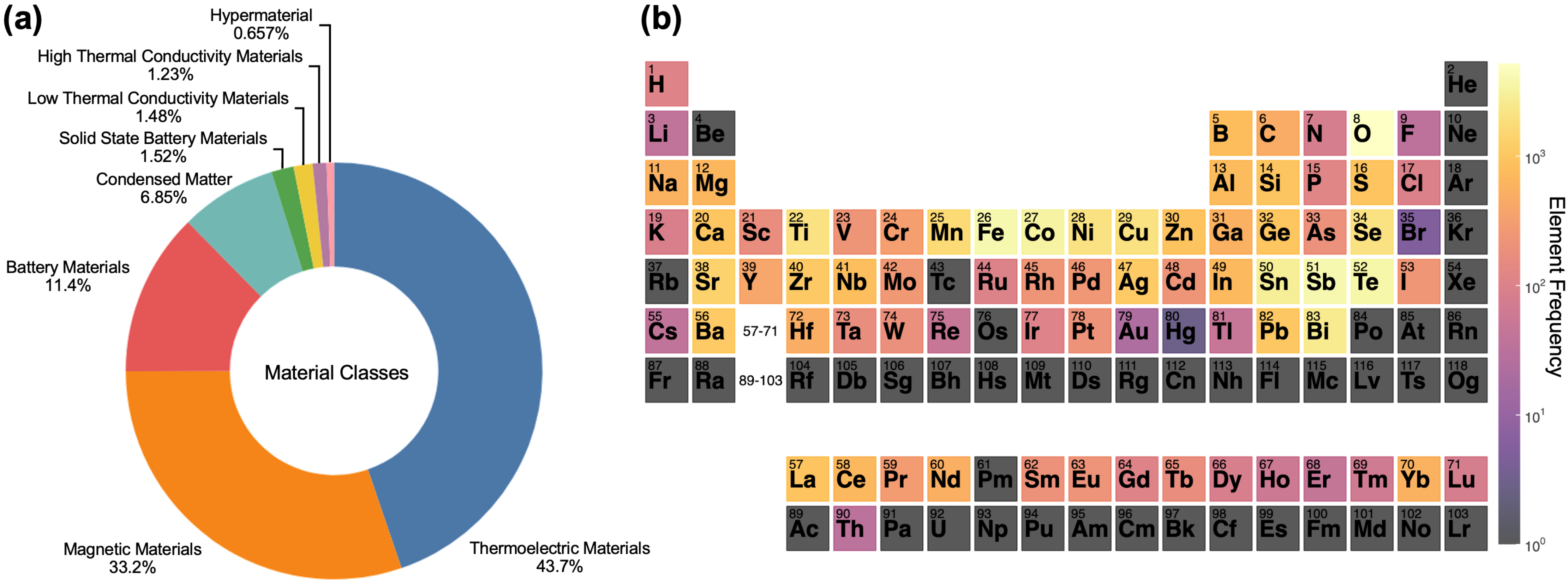}
  \caption{(a) Distribution of synthesis procedures in MatPROV by material type.
  (b) Periodic table visualization of the elemental frequency in materials included in MatPROV.}
  \label{fig:additional_data_analysis}
\end{figure}

\paragraph{Material Composition}

Figure~\ref{fig:additional_data_analysis}(a) illustrates the distribution of synthesis procedures in the MatPROV dataset, categorized by material type based on project classifications provided in Starrydata2.
The dataset primarily consists of thermoelectric materials (43.7\%), magnetic materials (33.2\%), and battery materials (11.4\%), with these three categories accounting for over 88\% of the dataset.
Figure~\ref{fig:additional_data_analysis}(b) presents a periodic table visualization showing the occurrence frequency of constituent elements in materials contained within the MatPROV dataset, based on their chemical compositions.
The results demonstrate that MatPROV covers a diverse range of elements, excluding artificial elements and noble gases.
However, a bias in elemental frequency is observed, reflecting the characteristics of the underlying source database, Starrydata2.
As a result, elements commonly associated with thermoelectric materials, such as Bi, Sb, Te, Pb, Se, and Sn, as well as transition metals typical of ferromagnetic materials, including Fe, Co, Ni, and Mn, exhibit high occurrence frequencies.

\paragraph{Synthesis Backbone}

We describe the method for constructing synthesis backbones shown in Figures~\ref{fig:data_analysis}(b) and 2(c).
To uncover the fundamental structure underlying material synthesis procedures, we constructed synthesis backbones from the activity nodes in our dataset as follows.
First, we analyzed all activity node labels and computed forward co-occurrence frequencies by aggregating all activity nodes that appear downstream from each node in the directed graph, regardless of whether they are directly connected.
This yielded a forward co-occurrence frequency matrix capturing how frequently one activity tends to follow another in synthesis procedures.
We then identified the most frequently co-occurring activity pair and used it as the starting point for backbone construction.
From this initial pair, we iteratively extended the backbone in both forward and backward directions along the directed edges by adding the most frequently co-occurring activity node at each step, as long as the frequency exceeded an empirically determined threshold of 25.
The resulting directed sequence of high-frequency activities, represented by green nodes and edges, constitutes the synthesis backbone.
In the backbone visualizations, edge weights represent co-occurrence frequencies.
Gray-colored activity nodes indicate candidate backbone nodes that surpassed the frequency threshold but were not selected because they were less frequent than the top candidate at that position.
This approach identifies dominant synthesis pathways and reveals a structural template characterizing material synthesis procedures in our dataset.

\subsection{Datasheets for Datasets}
\label{sec:appendix_datasheets}

The following questions were copied from ``Datasheets for Datasets''~\cite{gebru2021datasheets}.

\subsubsection{Motivation}

\textbf{For what purpose was the dataset created? Was there a specific task in mind? Was there a specific gap that needed to be filled? Please provide a description.}

MatPROV was developed to address the limitations of existing datasets on synthesis procedures, which often rely on fixed-field schemas or assume linear workflows.
By leveraging the PROV-DM standard, MatPROV enables graph-based representations that capture the structural complexity of real-world synthesis procedures.
This machine-interpretable format supports downstream applications, including automated synthesis planning, process optimization, and causal reasoning about process-property relationships.

\textbf{Who created this dataset (e.g., which team, research group) and on behalf of which entity (e.g., company, institution, organization)?}

MatPROV was created by SAKURA internet Inc.

\textbf{Who funded the creation of the dataset? (If there is an associated grant, please provide the name of the grantor and the grant name and number.)}

This work was not supported by any specific grant or funding agency.

\textbf{Any other comments?}

No.

\subsubsection{Composition}

\textbf{What do the instances that comprise the dataset represent (e.g., documents, photos, people, countries)? Are there multiple types of instances (e.g., movies, users, and ratings; people and interactions between them; nodes and edges)? Please provide a description.}

Instances in MatPROV represent material synthesis procedures extracted from scientific papers as PROV-DM-compliant directed graphs.
Each instance contains two types of nodes---entities (materials, including precursors, intermediate or final products, and experimental tools) and activities (experimental operations)---and edges representing usage and generation relationships.

\textbf{How many instances are there in total (of each type, if appropriate)?}

MatPROV comprises 2,367 synthesis procedures extracted from 1,568 scientific papers.

\textbf{Does the dataset contain all possible instances or is it a sample (not necessarily random) of instances from a larger set? If the dataset is a sample, then what is the larger set? Is the sample representative of the larger set (e.g., geographic coverage)? If so, please describe how this representativeness was validated/verified. If it is not representative of the larger set, please describe why not (e.g., to cover a more diverse range of instances, because instances were withheld or unavailable).}

MatPROV is a sample from a larger set of scientific literature, specifically limited to open-access papers curated by Starrydata2.

\textbf{What data does each instance consist of? ``Raw'' data (e.g., unprocessed text or images)or features? In either case, please provide a description.}

Each instance consists of structured synthesis procedures represented in PROV-JSONLD format, extracted from synthesis-related text in scientific papers using LLMs.

\textbf{Is there a label or target associated with each instance? If so, please provide a description.}

Each synthesis procedure includes a label field that encodes the material's chemical composition and key synthesis characteristics (e.g., ``CuGaTe2\_ball-milling'').
This label serves to identify and categorize procedures but is not a target for prediction tasks.

\textbf{Is any information missing from individual instances? If so, please provide a description, explaining why this information is missing (e.g., because it was unavailable). This does not include intentionally removed information, but might include, e.g., redacted text.}

Information may be missing due to limitations in the automated extraction process.
As shown in Section~\ref{sec:experiments}, the LLM-based extraction achieves imperfect accuracy, which can result in incomplete or partially missing nodes, edges, or parameters in the provenance graphs.

\textbf{Are relationships between individual instances made explicit (e.g., users' movie ratings, social network links)? If so, please describe how these relationships are made explicit.}

No.
Each synthesis procedure is treated as an independent instance, though procedures may be related through common materials, operations, or source papers.

\textbf{Are there recommended data splits (e.g., training, development/validation, testing)? If so, please provide a description of these splits, explaining the rationale behind them.}

No specific data splits are recommended.
However, a set of 30 papers with 44 synthesis procedures has been manually annotated by a domain expert to serve as ground truth.
This subset can be used as a test set for evaluation purposes, as demonstrated in Section~\ref{sec:experiments}.

\textbf{Are there any errors, sources of noise, or redundancies in the dataset? If so, please provide a description.}

As the synthesis procedures were extracted using LLMs, errors and inaccuracies may be present.
Evaluation shows imperfect extraction accuracy, meaning several synthesis procedures may contain incorrect nodes, edges, or parameters.
Additionally, 4 procedures (0.2\%) resulted in cyclic graphs, and 41 (1.7\%) included isolated nodes, both of which violate the expected directed acyclic graph structure.

\textbf{Is the dataset self-contained, or does it link to or otherwise rely on external resources (e.g., websites, tweets, other datasets)? If it links to or relies on external resources, a) are there guarantees that they will exist, and remain constant, over time; b) are there official archival versions of the complete dataset (i.e., including the external resources as they existed at the time the dataset was created); c) are there any restrictions (e.g., licenses, fees) associated with any of the external resources that might apply to a future user? Please provide descriptions of all external resources and any restrictions associated with them, as well as links or other access points, as appropriate.}

The dataset includes links between extracted synthesis procedures and their corresponding paper DOIs, enabling users to trace back to the original publications.
While the dataset relies on the continued availability of these scientific papers, all source papers are open access.
The dataset is self-contained regarding the extracted synthesis procedures, which are fully included in PROV-JSONLD format.

\textbf{Does the dataset contain data that might be considered confidential (e.g., data that is protected by legal privilege or by doctor-patient confidentiality, data that includes the content of individuals' non-public communications)? If so, please provide a description.}

No.

\textbf{Does the dataset contain data that, if viewed directly, might be offensive, insulting, threatening, or might otherwise cause anxiety? If so, please describe why.}

No.

\textbf{Does the dataset relate to people? If not, you may skip the remaining questions in this section.}

No.

\textbf{Does the dataset identify any subpopulations (e.g., by age, gender)? If so, please describe how these subpopulations are identified and provide a description of their respective distributions within the dataset.}

N/A.

\textbf{Is it possible to identify individuals (i.e., one or more natural persons), either directly or indirectly (i.e., in combination with other data) from the dataset? If so, please describe how.}

N/A.

\textbf{Does the dataset contain data that might be considered sensitive in any way (e.g., data that reveals racial or ethnic origins, sexual orientations, religious beliefs, political opinions or union memberships, or locations; financial or health data; biometric or genetic data; forms of government identification, such as social security numbers; criminal history)? If so, please provide a description.}

N/A.

\textbf{Any other comments?}

No.

\subsubsection{Collection Process}

\textbf{How was the data associated with each instance acquired? Was the data directly observable (e.g., raw text, movie ratings), reported by subjects (e.g., survey responses), or indirectly inferred/derived from other data (e.g., part-of-speech tags, model-based guesses for age or language)? If data was reported by subjects or indirectly inferred/derived from other data, was the data validated/verified? If so, please describe how.}

Data were indirectly derived from scientific papers through a multi-step automated extraction process using LLMs.
Synthesis-relevant text was first identified and extracted, and then converted into structured synthesis procedure graphs in PROV-JSONLD format.

\textbf{What mechanisms or procedures were used to collect the data (e.g., hardware apparatus or sensor, manual human curation, software program, software API)? How were these mechanisms or procedures validated?}

The collection process involved three main steps: (1) converting scientific paper PDFs to structured XML using GROBID v0.8.2, (2) extracting synthesis-relevant text using OpenAI’s GPT-4o mini, and (3) extracting synthesis procedures in PROV-JSONLD format using the OpenAI API.
Validation was performed by comparing the extracted graphs against expert-annotated ground truth using precision, recall, and F1-score metrics at both the structural and parametric levels, as described in Section~\ref{sec:experiments}.

\textbf{If the dataset is a sample from a larger set, what was the sampling strategy (e.g., deterministic, probabilistic with specific sampling probabilities)?}

MatPROV contains synthesis procedures only from open-access papers curated by Starrydata2, to ensure copyright compliance.

\textbf{Who was involved in the data collection process (e.g., students, crowdworkers, contractors) and how were they compensated (e.g., how much were crowdworkers paid)?}

All authors of the paper were involved in the data collection process.

\textbf{Over what timeframe was the data collected? Does this timeframe match the creation timeframe of the data associated with the instances (e.g., recent crawl of old news articles)? If not, please describe the timeframe in which the data associated with the instances was created.}

MatPROV was created in July 2025 based on a corpus of scientific papers published between 1982 and 2025.

\textbf{Were any ethical review processes conducted (e.g., by an institutional review board)? If so, please provide a description of these review processes, including the outcomes, as well as a link or other access point to any supporting documentation.}

No specific ethical review processes are mentioned in the paper, as the work involves analysis of publicly available scientific literature rather than human subjects research.

\textbf{Does the dataset relate to people? If not, you may skip the remaining questions in this section.}

No.

\textbf{Did you collect the data from the individuals in question directly, or obtain it via third parties or other sources (e.g., websites)?}

N/A.

\textbf{Were the individuals in question notified about the data collection? If so, please describe (or show with screenshots or other information) how notice was provided, and provide a link or other access point to, or otherwise reproduce, the exact language of the notification itself.}

N/A.

\textbf{Did the individuals in question consent to the collection and use of their data? If so, please describe (or show with screenshots or other information) how consent was requested and provided, and provide a link or other access point to, or otherwise reproduce, the exact language to which the individuals consented.}

N/A.

\textbf{If consent was obtained, were the consenting individuals provided with a mechanism to revoke their consent in the future or for certain uses? If so, please provide a description, as well as a link or other access point to the mechanism (if appropriate).}

N/A.

\textbf{Has an analysis of the potential impact of the dataset and its use on data subjects (e.g., a data protection impact analysis) been conducted? If so, please provide a description of this analysis, including the outcomes, as well as a link or other access point to any supporting documentation.}

N/A.

\textbf{Any other comments?}

No.

\subsubsection{Preprocessing/cleaning/labeling}

\textbf{Was any preprocessing/cleaning/labeling of the data done (e.g., discretization or bucketing, tokenization, part-of-speech tagging, SIFT feature extraction, removal of instances, processing of missing values)? If so, please provide a description. If not, you may skip the remainder of the questions in this section.}

Yes.
Extensive preprocessing was performed, including: (1) PDF to structured XML conversion using GROBID, (2) extracting main body text excluding titles, abstracts, and references, (3) identification and extraction of synthesis-relevant text using LLMs, (4) converting synthesis text to PROV-JSONLD format, and (5) post-processing to add standardized metadata fields.

\textbf{Was the "raw" data saved in addition to the preprocessed/cleaned/labeled data (e.g., to support unanticipated future uses)? If so, please provide a link or other access point to the "raw" data.}

The raw data, that is, the text of the source papers, was not redistributed as part of the dataset.
However, MatPROV includes DOI links for all extracted synthesis procedures, enabling users to access the original publications.
Since all source papers are open-access, the raw text remains publicly accessible.

\textbf{Is the software used to preprocess/clean/label the instances available? If so, please provide a link or other access point.}

Yes.
Code is publicly available at \url{https://github.com/MatPROV-project/matprov-experiments}.

\textbf{Any other comments?}

No.

\subsubsection{Uses}

\textbf{Has the dataset been used for any tasks already? If so, please provide a description.}

MatPROV was not used for downstream tasks in this study.
Instead, it was analyzed in Section~\ref{sec:dataset_analysis} to examine the distribution of graph sizes and material-type-specific synthesis patterns.

\textbf{Is there a repository that links to any or all papers or systems that use the dataset? If so, please provide a link or other access point.}

No.

\textbf{What (other) tasks could the dataset be used for?}

The dataset could potentially be used for tasks such as automated synthesis planning, process optimization, understanding relationships between synthesis conditions and material properties, and benchmarking graph extraction from scientific texts.

\textbf{Is there anything about the composition of the dataset or the way it was collected and preprocessed/cleaned/labeled that might impact future uses? For example, is there anything that a future user might need to know to avoid uses that could result in unfair treatment of individuals or groups (e.g., stereotyping, quality of service issues) or other undesirable harms (e.g., financial harms, legal risks) If so, please provide a description. Is there anything a future user could do to mitigate these undesirable harms?}

Yes.
The automated extraction process introduces potential errors that could lead to incorrect synthesis information if used without validation.
This may result in failed experiments, wasted resources, or safety hazards in laboratory settings.
Users are strongly advised to validate critical synthesis procedures against the original papers before practical implementation, as discussed in Appendix~\ref{sec:negative_impacts}.
Additionally, the dataset exhibits a bias toward certain material types---particularly thermoelectric and magnetic materials---due to the composition of the source corpus, which may limit its generalizability to other domains.

\textbf{Are there tasks for which the dataset should not be used? If so, please provide a description.}

MatPROV should not be used for direct synthesis implementation without validation against original sources due to extraction errors.

\textbf{Any other comments?}

No.

\subsubsection{Distribution}

\textbf{Will the dataset be distributed to third parties outside of the entity (e.g., company, institution, organization) on behalf of which the dataset was created? If so, please provide a description.}

Yes.
MatPROV is publicly available.

\textbf{How will the dataset will be distributed (e.g., tarball on website, API, GitHub)? Does the dataset have a digital object identifier (DOI)?}

We released MatPROV on the Hugging Face Hub with DOI: \url{https://doi.org/10.57967/hf/6382}.

\textbf{When will the dataset be distributed?}

MatPROV has already been released.

\textbf{Will the dataset be distributed under a copyright or other intellectual property (IP) license, and/or under applicable terms of use (ToU)? If so, please describe this license and/or ToU, and provide a link or other access point to, or otherwise reproduce, any relevant licensing terms or ToU, as well as any fees associated with these restrictions.}

MatPROV is released under the CC BY 4.0 license.

\textbf{Have any third parties imposed IP-based or other restrictions on the data associated with the instances? If so, please describe these restrictions, and provide a link or other access point to, or otherwise reproduce, any relevant licensing terms, as well as any fees associated with these restrictions.}

No.

\textbf{Do any export controls or other regulatory restrictions apply to the dataset or to individual instances? If so, please describe these restrictions, and provide a link or other access point to, or otherwise reproduce, any supporting documentation.}

No.

\textbf{Any other comments?}

No.

\subsubsection{Maintenance}

\textbf{Who is supporting/hosting/maintaining the dataset?}

The datasets and code used to construct the datasets are hosted on the Hugging Face Hub and GitHub, respectively, to ensure high availability and long-term preservation.
The authors are responsible for maintaining these resources and address issues or updates as needed.

\textbf{How can the owner/curator/manager of the dataset be contacted (e.g., email address)?}

Please contact hi-tsuruta@sakura.ad.jp.

\textbf{Is there an erratum? If so, please provide a link or other access point.}

There are no errata for our initial release.
Errata will be published on the Hugging Face Hub and GitHub when needed.

\textbf{Will the dataset be updated (e.g., to correct labeling errors, add new instances, delete instances')? If so, please describe how often, by whom, and how updates will be communicated to users (e.g., mailing list, GitHub)?}

If we find any issues with our datasets and update them, we will release an updated version on the Hugging Face Hub.

\textbf{If the dataset relates to people, are there applicable limits on the retention of the data associated with the instances (e.g., were individuals in question told that their data would be retained for a fixed period of time and then deleted)? If so, please describe these limits and explain how they will be enforced.}

N/A.

\textbf{Will older versions of the dataset continue to be supported/hosted/maintained? If so, please describe how. If not, please describe how its obsolescence will be communicated to users.}

Yes.
If we plan to update the datasets, we will maintain the old version and then release the updated version.

\textbf{If others want to extend/augment/build on/contribute to the dataset, is there a mechanism for them to do so? If so, please provide a description. Will these contributions be validated/verified? If so, please describe how. If not, why not? Is there a process for communicating/distributing these contributions to other users? If so, please provide a description.}

We welcome and encourage others to extend/augment/build on/contribute to the datasets.
If others would like to contribute to our datasets, they can submit a pull request on GitHub or contact us via email.

\textbf{Any other comments?}

No.


\begin{thebibliography}{10}
\providecommand{\url}[1]{\texttt{#1}}
\providecommand{\urlprefix}{URL }
\providecommand{\doi}[1]{https://doi.org/#1}

\bibitem{GROBID}
{GROBID}. \url{https://github.com/kermitt2/grobid} (2008--2025)

\bibitem{ali2021review}
Ali, A., Shah, T., Ullah, R., Zhou, P., Guo, M., Ovais, M., Tan, Z., Rui, Y.:
  Review on recent progress in magnetic nanoparticles: Synthesis,
  characterization, and diverse applications. Frontiers in Chemistry
  \textbf{9},  629054 (2021)

\bibitem{baig2021nanomaterials}
Baig, N., Kammakakam, I., Falath, W.: Nanomaterials: A review of synthesis
  methods, properties, recent progress, and challenges. Materials Advances
  \textbf{2},  1821--1871 (2021)

\bibitem{belhajjame2013prov}
Belhajjame, K., B’Far, R., Cheney, J., Coppens, S., Cresswell, S., Gil, Y.,
  Groth, P., Klyne, G., Lebo, T., McCusker, J., et~al.: {PROV}-{DM}: {T}he
  {PROV} {D}ata {M}odel. W3C Recommendation  (2013)

\bibitem{boolchandani2018preparation}
Boolchandani, S., Srivastava, S., Vijay, Y.: Preparation of {I}n{S}e thin films
  by thermal evaporation method and their characterization: structural,
  optical, and thermoelectrical properties. Journal of Nanotechnology
  \textbf{2018},  9380573 (2018)

\bibitem{d2023thermoelectric}
d’Angelo, M., Galassi, C., Lecis, N.: Thermoelectric materials and
  applications: a review. Energies  \textbf{16}(17), ~6409 (2023)

\bibitem{fu2016enhancing}
Fu, C., Wu, H., Liu, Y., He, J., Zhao, X., Zhu, T.: Enhancing the figure of
  merit of heavy-band thermoelectric materials through hierarchical phonon
  scattering. Advanced Science  \textbf{3}(8),  1600035 (2016)

\bibitem{gebru2021datasheets}
Gebru, T., Morgenstern, J., Vecchione, B., Vaughan, J.W., Wallach, H., Iii,
  H.D., Crawford, K.: Datasheets for datasets. Communications of the ACM
  \textbf{64}(12),  86--92 (2021)

\bibitem{he2015mossbauer}
He, Y., Yang, X., Lin, J., Lin, Q., Dong, J.: M{\"o}ssbauer spectroscopy,
  structural and magnetic studies of {Z}n$^{2+}$ substituted magnesium ferrite
  nanomaterials prepared by sol-gel method. Journal of Nanomaterials
  \textbf{2015},  854840 (2015)

\bibitem{huynh2016prov}
Huynh, T.D., Michaelides, D.T., Moreau, L.: {PROV}-{JSONLD}: A {JSON} and
  linked data representation for provenance. In: International Provenance and
  Annotation Workshop. pp. 173--177. Springer (2016)

\bibitem{jiang2025applications}
Jiang, X., Wang, W., Tian, S., Wang, H., Lookman, T., Su, Y.: Applications of
  natural language processing and large language models in materials discovery.
  npj Computational Materials  \textbf{11}, ~79 (2025)

\bibitem{katsura2019data}
Katsura, Y., Kumagai, M., Kodani, T., Kaneshige, M., Ando, Y., Gunji, S., Imai,
  Y., Ouchi, H., Tobita, K., Kimura, K., et~al.: Data-driven analysis of
  electron relaxation times in {P}b{T}e-type thermoelectric materials. Science
  and Technology of Advanced Materials  \textbf{20}(1),  511--520 (2019)

\bibitem{katsura2025starrydata}
Katsura, Y., Kumagai, M., Mato, T., Takada, Y., Ando, Y., Fujita, E., Hosono,
  F., Koyama, E., Mudasar, F., Phuong, T.N.T., et~al.: Starrydata: from
  published plots to shared materials data. Science and Technology of Advanced
  Materials: Methods  \textbf{5}(1),  2506976 (2025)

\bibitem{kononova2019text}
Kononova, O., Huo, H., He, T., Rong, Z., Botari, T., Sun, W., Tshitoyan, V.,
  Ceder, G.: Text-mined dataset of inorganic materials synthesis recipes.
  Scientific data  \textbf{6}, ~203 (2019)

\bibitem{manley2019intrinsic}
Manley, M.E., Hellman, O., Shulumba, N., May, A.F., Stonaha, P.J., Lynn, J.W.,
  Garlea, V.O., Alatas, A., Hermann, R.P., Budai, J.D., et~al.: Intrinsic
  anharmonic localization in thermoelectric {P}b{S}e. Nature communications
  \textbf{10}, ~1928 (2019)

\bibitem{mao2020decoupling}
Mao, T., Qiu, P., Hu, P., Du, X., Zhao, K., Wei, T.R., Xiao, J., Shi, X., Chen,
  L.: Decoupling thermoelectric performance and stability in liquid-like
  thermoelectric materials. Advanced Science  \textbf{7}(1),  1901598 (2020)

\bibitem{moreau2020prov}
Moreau, L., Huynh, T.D.: The {PROV}-{JSONLD} serialization: A {JSON}-{LD}
  representation for the {PROV} data model. In: International Provenance and
  Annotation Workshop. pp. 51--67. Springer (2020)

\bibitem{przybyl2023magnetic}
Przyby{\l}, A., Wnuk, I., Wys{\l}ocki, J., Kutynia, K., Ka{\'z}mierczak, M.,
  Rychta, M., G{\k{e}}bara, P.: Magnetic interactions and coercivity mechanism
  in nanocrystalline {N}d--{F}e--{B} ribbons with {N}b addition. Acta Physica
  Polonica A  \textbf{144}(5), ~333 (2023)

\bibitem{schilling2025text}
Schilling-Wilhelmi, M., R{\'\i}os-Garc{\'\i}a, M., Shabih, S., Gil, M.V.,
  Miret, S., Koch, C.T., M{\'a}rquez, J.A., Jablonka, K.M.: From text to
  insight: large language models for chemical data extraction. Chemical Society
  Reviews  \textbf{54},  1125--1150 (2025)

\bibitem{shaba2021critical}
Shaba, E.Y., Jacob, J.O., Tijani, J.O., Suleiman, M.A.T.: A critical review of
  synthesis parameters affecting the properties of zinc oxide nanoparticle and
  its application in wastewater treatment. Applied Water Science  \textbf{11},
  ~48 (2021)

\bibitem{shi2024llm}
Shi, L., Liu, Z., Yang, Y., Wu, W., Zhang, Y., Zhang, H., Lin, J., Wu, S.,
  Chen, Z., Li, R., et~al.: {LLM}-based {MOF}s synthesis condition extraction
  using few-shot demonstrations. arXiv preprint arXiv:2408.04665  (2024)

\bibitem{walker2023extracting}
Walker, N., Lee, S., Dagdelen, J., Cruse, K., Gleason, S., Dunn, A., Ceder, G.,
  Alivisatos, A.P., Persson, K.A., Jain, A.: Extracting structured
  seed-mediated gold nanorod growth procedures from scientific text with
  {LLM}s. Digital Discovery  \textbf{2},  1768--1782 (2023)

\bibitem{wang2022dataset}
Wang, Z., Kononova, O., Cruse, K., He, T., Huo, H., Fei, Y., Zeng, Y., Sun, Y.,
  Cai, Z., Sun, W., et~al.: Dataset of solution-based inorganic materials
  synthesis procedures extracted from the scientific literature. Scientific
  data  \textbf{9}, ~231 (2022)

\bibitem{wei2014tau}
Wei, J., Song, Z., Yang, Y., Liu, S., Du, H., Han, J., Zhou, D., Wang, C.,
  Yang, Y., Franz, A., et~al.: $\tau$-{M}n{A}l with high coercivity and
  saturation magnetization. AIP Advances  \textbf{4}(12),  127113 (2014)

\bibitem{zheng2023chatgpt}
Zheng, Z., Zhang, O., Borgs, C., Chayes, J.T., Yaghi, O.M.: Chat{GPT} chemistry
  assistant for text mining and the prediction of {MOF} synthesis. Journal of
  the American Chemical Society  \textbf{145}(32),  18048--18062 (2023)

\end{thebibliography}
\end{document}